\documentclass[11pt,a4paper]{article}
\usepackage[hyperref]{emnlp-ijcnlp-2019}
\usepackage{latexsym}

\usepackage{graphicx}
\usepackage{amsmath}
\usepackage{booktabs}
\usepackage{algorithm}
\usepackage{algorithmic}
\usepackage{url}
\usepackage{subfigure}
\usepackage{color}

\definecolor{shadecolor}{gray}{0.95}
\newcommand{\algshade}[1]{
\hspace*{-\fboxsep}
\colorbox{shadecolor}{
\parbox{\linewidth}{
#1
}}
}

\newcommand{\csize}{
\fontsize{8}{8}\selectfont
}

\newcommand{\tabsize}{
\fontsize{7}{7}\selectfont
}

\renewcommand\algorithmiccomment[1]{
  {
  	{
  	{\textit{\%#1}}
  	}
  }
}

\newcommand{\cI}{{\mathcal{I}}}

\newcommand{\cH}{{\mathcal{H}}}

\newcommand{\bs}[1]{\boldsymbol{#1}}

\newcommand{\tabref}[1]{Tab.~\ref{#1}}
\newcommand{\figref}[1]{Fig.~\ref{#1}}

\newcommand{\secref}[1]{Sec.~\ref{#1}}

\newcommand{\algoref}[1]{Alg.~\ref{#1}}

\aclfinalcopy % Uncomment this line for the final submission

\title{Nearly-Unsupervised Hashcode Representations\\ 
for Relation Extraction}

\author{
Sahil Garg\\
USC ISI\\
  {\tt sahilgar@usc.edu} \\
  \And
  Aram Galstyan\\
USC ISI\\
  {\tt galstyan@isi.edu} \\
  \And
  Greg Ver Steeg\\
USC ISI\\
  {\tt gregv@isi.edu} \\
  \And
  Guillermo Cecchi\\
  IBM Research, NY\\
  {\tt gcecchi@us.ibm.com} \\
}

\date{}

\begin{document}
\maketitle

\begin{abstract}
Recently, 
\emph{kernelized locality sensitive hashcodes} have been successfully employed as representations of natural language text, 
% (parse trees, shortest paths, text sentences, etc.), 
especially showing high relevance to biomedical relation extraction tasks.
% 
% Owing to the inherent \emph{computationally scalable} nature of the hashcodes based approach, and its theoretical properties due to locality sensitivity of hash functions, 
% 
% In this paper, 
% 
% In their work, hashcode representations are optimized in a supervised manner which are then fed into a supervised Random Forest classifier.
% 
In this paper, we propose to optimize the hashcode representations in a \emph{nearly unsupervised} manner, in which we only use data points, but not their class labels, for learning. %Besides the data points coming from a training set, this setting requires data points either from a test set, or one can treat a random subset of a training set itself as a pseudo-test set. 
% 
% In the latter case, hashcodes are learned in a purely unsupervised manner, whereas the former case corresponds to nearly-unsupervised learning.
% 
The optimized hashcode representations are then fed to a supervised classifier following the prior work.
% 
% using data points from training set, as well as data points from a test set or a pseudo-test set
% 
% We extend their approach with a \emph{semi-supervised} learning framework, using only the information on which set a natural language data point belongs to, a training set or test set~(or a pseudo-test set), whereas the actual class labels of examples are fed as input only to the final supervised classifier, such as a Random Forest, along with the optimized hashcodes of the examples~(representation vectors).
% 
% for \emph{inductive} as well as \emph{transductive} settings, 
% in which we 
% 
% constructing hash functions one by one with our \emph{information-theoretic greedy} algorithm.
% 
% , using only the information on which set a natural language data point belongs to, training or test set, whereas the actual class labels of examples are fed as input only to the final supervised classifier, such as a Random Forest, along with the optimized hashcodes of the examples~(representation vectors). 
% 
% This kind of \emph{nearly unsupervised} representation learning paradigm 
% This representation learning approach
% 
This nearly unsupervised approach allows fine-grained optimization of each hash function, 
% in contrast to the previous supervised hashcode learning approach that optimizes only the parameters which are shared amongst the hash functions, 
which is particularly suitable for building hashcode representations generalizing from a training set to a test set.
% 
% while being \emph{robust} to overfitting to a train set. 
% 
% One of the key ideas in our approach is of learning \emph{hierarchical} hashcodes by constructing a hash function from examples \emph{sampled locally} from one of the \emph{clusters} defined as per a subset of the previously optimized hash functions, so as to capture fine-grained differences between examples.
% 
% \todo{summary on ideas}
% 
We empirically evaluate the proposed approach for biomedical relation extraction tasks, 
% from the previous work using four public datasets,
obtaining significant accuracy improvements w.r.t. state-of-the-art supervised and semi-supervised approaches.
% 
% The applicability of our \emph{generic} framework is also demonstrated, beyond kernel similarity functions, to \emph{neural networks}.

\end{abstract}

% \vspace{-2mm}
\section{Introduction}
\begin{figure*}[tp!]
\centering
\includegraphics[
width=1.3\columnwidth]{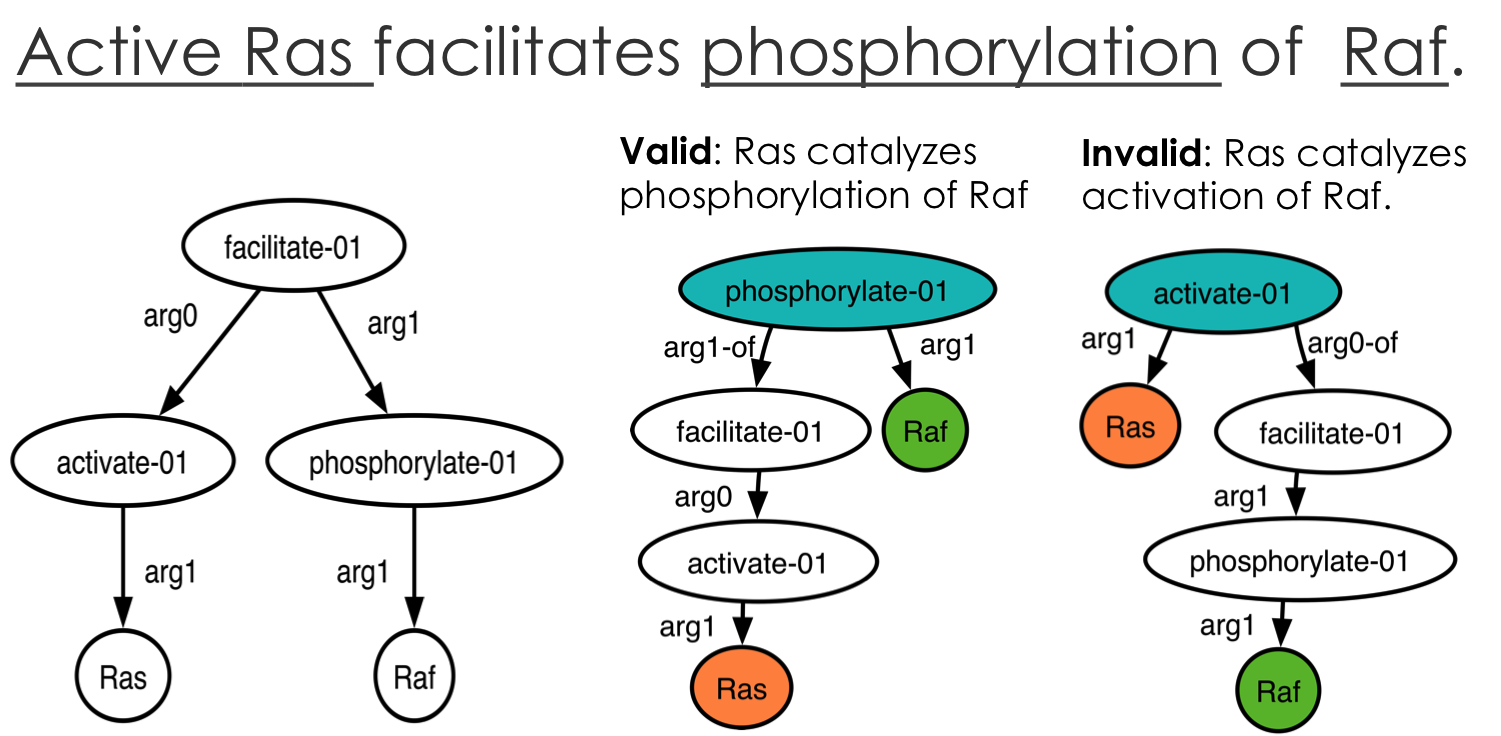}
% 
% \vspace{-4.75mm}
% 
\caption{
% \csizeten
% 
% In this figure, we describe a task of biomedical relation extraction.
% 
On the left, we show an abstract meaning representation~(AMR) of a sentence.
% 
% , and in the sentence, the tokens corresponding to bio-entities 
% % (proteins, chemicals,
% etc.) 
% or interaction types are underlined. 
% 
As per the semantics of the sentence, there is a valid biomedical relationship between the two proteins, Ras and Raf, i.e. Ras catalyzes phosphorylation of Raf; the relation corresponds to a subgraph extracted from the AMR.
% , shown with color-coding for the constituents: an interaction type (green) and two-three bio-entities either participating in the interaction (parrot), or catalyzing it (orange).
% 
On the other hand, one of the many invalid biomedical relationships that one could infer is, Ras catalyzes activation of Raf, for which we show the corresponding subgraph too. A given candidate relation automatically hypothesized from the sentence, is binary classified, as valid or invalid, using the subgraph as features.
% , as done in previous works.
}
% \vspace{-4mm}
\label{fig:ee}
\end{figure*}

% \vspace{-1mm}
% 
% 
In natural language processing, one important but a highly challenging task is of identifying biological entities and their relations from biomedical text, as illustrated in \figref{fig:ee}, relevant for real world problems, such as personalized cancer treatments~\cite{rzhetsky2016big,hahn2015domain,cohen2015darpa}. In the previous works, the task of biomedical relation extraction is formulated as of binary classification of natural language structures; one of the primary challenges to solve the problem is that the number of data points annotated with class labels (in a training set) is small due to high cost of annotations by biomedical domain experts, and further, bio-text sentences in a test set can vary significantly w.r.t. the ones from a training set due to practical aspects, such as high diversity of research topics or writing styles, hedging, etc.
Considering such challenges for the task, many classification models based on kernel similarity functions have been proposed~\cite{garg2016extracting,chang2016pipe,tikk2010comprehensive,miwa2009protein,airola2008all,mooney2005subsequence}, and recently, many neural networks based classification models have also been explored~\cite{kavuluru2017extracting,peng2017deep,hsieh2017identifying,raobiomedical,nguyen2015relation}, including the ones doing adversarial learning using the knowledge of data points~(excluding class labels) from a test set, or semi-supervised variational autoencoders~\cite{rios2018generalizing,ganin2016domain,zhang2019exploring,kingma14semisupervised}.
        
In a very recent work, kernelized locality sensitive hashcodes based representation learning approach has been proposed that has shown to be the most successful in terms of accuracy and computational efficiency for the task~\cite{garg2019kernelized}.
% ,garg2018stochastic,joly2011random,kulis2009kernelized}.
% 
% 
The model parameters, shared between all the hash functions, are optimized in a supervised manner, whereas an individual hash function is constructed in a \emph{randomized} fashion.
% % 
% % , for a fixed kernel compute cost, where each hash function can 
% serving for randomized semantic categorization of data points~(binary semantic feature).
% % 
% 
% In their approach, kernelized~(binary) hashcode of a data point acts as its representation vector, computed as a function of kernel similarities of the example w.r.t. (only) a constant sized \emph{reference set} of data points, and then the representations vectors are fed into a Random Forest of decision trees as the final classification model.
% 
% The choice of the reference set and the parameters of a kernel function are optimized in a purely supervised manner, shared between all the (kernel based) hash functions, whereas an individual hash function is constructed in a \emph{randomized} fashion,
% % 
% % , for a fixed kernel compute cost, where each hash function can 
% serving for randomized semantic categorization of data points~(binary semantic feature).
% 
% , and the bit value for an example should mean presence or absence of language patterns in the example. 
% 
The authors suggest to obtain thousands of (randomized) semantic features extracted from natural language data points into binary hashcodes, and then making classification decision as per the \emph{features} using hundreds of decision trees, which is the core of their robust classification approach. 
% 
% Although hashcode representation learning is a very promising direction of research, there is a significant scope for improving upon the supervised approach presented in the previous work. 
% 
Even if we extract thousands of semantic features using the hashing approach, it is difficult to ensure that the features extracted from training data points would generalize to a test set. While the inherent randomness in constructing hash functions from a training set can help towards generalization in the case of absence of a test set, there should be better alternatives if we do have the knowledge of a test set of data points, or a subset of a training set treated as a \emph{pseudo-test set}. What if we construct hash functions in an intelligent manner via exploiting the additional knowledge of unlabeled data points in a test/pseudo-test set, performing \emph{fine-grained optimization of each hash function} rather than relying upon randomness, so as to extract semantic features 
which 
% are \emph{not purely random} but 
generalize?
%
%In relation to the previous works, their model is more robust w.r.t. overfitting a training dataset, relying upon basic ideas, such as generating a large number of hash functions so as to obtain a large number of features, and building a decision tree in RF using only a small subset of features.
%
% From this perspective, we observe two limitations in the previous hashcode representation learning approach.
% 
% First limitation is that their approach is limited to supervised settings.
% 
% Second limitation is that the proposed supervised learning of hash functions in their work allows optimizing only those aspects of the model which are common or global to all the hash functions, such as the choice of a reference set or kernel parameters, while keeping fine-grained aspects of a hash function randomized. 

\begin{figure*}[tp!]
\centering
\subfigure[
% \tabsize
Hash function choices
]{
\includegraphics[
width=0.7\columnwidth]{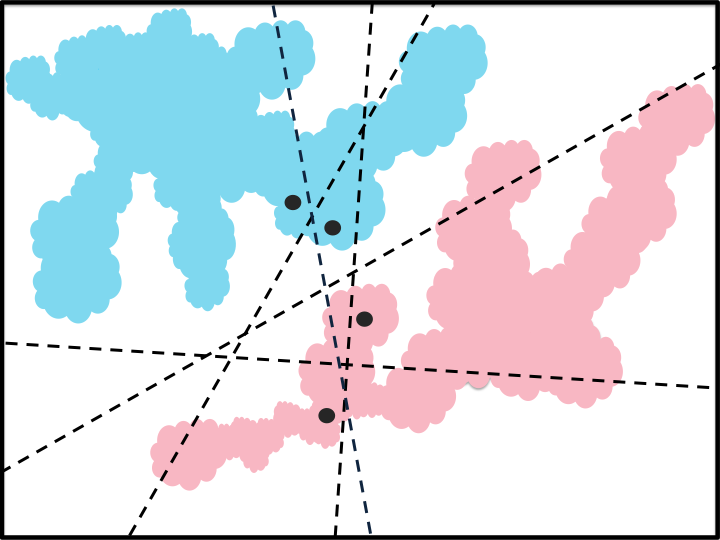}
\label{fig:hash_choices}
}
\hspace{10.75mm}
\subfigure[
% \tabsize
Generalizing hash function]{
\includegraphics[
width=0.7\columnwidth]{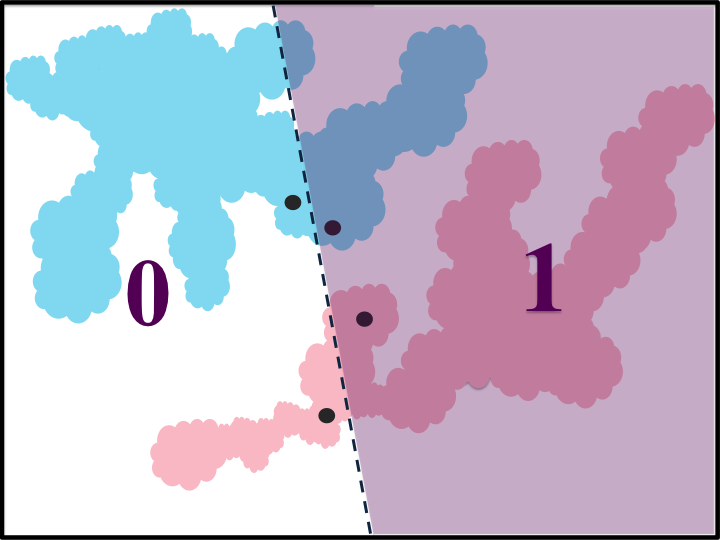}
\label{fig:hash_choice_optimal}
}
% 
% % \hspace{6mm}
% \hspace{-2mm}
% % 
% \subfigure[
% \tabsize
% Generalizing hash functions
% % LSH functions respecting train-test data distribution
% ]{
% \includegraphics[
% width=0.48\columnwidth]{desired.png}
% \label{fig:not_agnostic}}
% % 
% \hspace{-2mm}
% % 
% \subfigure[
% \tabsize
% Non-generalizing hash functions
% % LSH functions agnostic to train-test data distribution
% ]{
% \includegraphics[
% width=0.48\columnwidth]{undesired.png}
% \label{fig:agnostic}
% }
%
% 
% \vspace{-3.75mm}
% 
\caption{
% \csizeten
% 
In this figure, we illustrate how to construct hash functions which generalize across a training and a test set.
Blue and red colors denote training and test data points respectively. A hash function is constructed by splitting a very small subset of data points into two parts. In \figref{fig:hash_choices}, given the set of four data points selected randomly, there are many choices possible for splitting the set, corresponding to difference choices of hash functions which are denoted with dashed lines. The optimal choice of hash function is shown in \figref{fig:hash_choice_optimal} since it generalizes well, assigning same value to many data points from both training \& test sets.
}
% \vspace{-3mm}
\label{fig:klsh_ssh_concepts}
\end{figure*}

Along these lines, we propose a new framework for learning hashcode representations accomplishing two important (inter-related) extensions w.r.t. the previous work:
    
(a) We propose to use a \emph{nearly unsupervised} hashcode representation learning setting, in which we use only the knowledge of which set a data point comes from, a training set or a test/pseudo-test set, along with the data point itself, whereas the actual class labels of data points from a training set are input only to the final supervised-classifier, such as a Random Forest, which takes input of the learned hashcodes as representation~(feature) vectors of data points along with their class labels;
% 
% using either the knowledge of examples from a test set or a large set of unlabeled examples, so as to explicitly maximize generalization of hashcode representations from a train set to a test set; 

(b) We introduce multiple concepts for \emph{fine-grained~(discrete) optimization} of hash functions, employed in our novel \emph{information-theoretic algorithm} that constructs hash functions greedily one by one. In supervised settings, fine-grained (greedy) optimization of hash functions could lead to overfitting whereas, in our proposed nearly-unsupervised framework, it allows \emph{flexibility for explicitly maximizing the generalization capabilities} of hash functions.

For a task of biomedical relation extraction, we evaluate our approach on four public datasets, and obtain significant gains in F1 scores w.r.t. state-of-the-art models including kernel-based approaches as well the ones based on semi-supervised learning of neural networks. We also show how to employ our framework for learning locality sensitive hashcode representations using neural networks.\footnote{Code: \url{https://github.com/sgarg87/nearly_unsupervised_hashcode_representations}}

% \vspace{-2mm}
\section{Problem Formulation \& Background}
% \vspace{-1mm}
% 
\label{sec:ps}

% In this section, we describe the approach of kernelized hashcode representations~\cite{garg2019kernelized} which has shown state-of-the-art results for bio-medical relation extraction task though the approach, in principle, is applicable for any NLP task involving classification of natural language structures.

% \paragraph{Biomedical relation extraction\\}
% 
In \figref{fig:ee}, we demonstrate how biomedical relations between entities
% , like proteins, molecules, chemicals, 
are extracted from the semantic~(or syntactic) parse of a sentence. As we see, the task is formulated as of binary classification of natural language sub-structures extracted from the semantic parse.
% For the purpose of understanding the basics of hashcode representation approach, from here on, we consider a general problem setting as below.
% 
% 
% \paragraph{Problem settings\\}
% 
Suppose we have natural language structures, $\bs{S} = \{S_i\}_1^N$, such as parse trees, shortest paths, text sentences, etc, with corresponding class labels, $\bs{y} = \{y_i\}_1^N$. 
For the data points coming from a training set and a test set, we use notations, $\bs{S}_T$, and $\bs{S}_*$, respectively; same applies for the class labels.
In addition, we define indicator variable, $\bs{x} = \{x_i\}_{i=1}^N$, for $\bs{S}$, with $x_i \in \{0, 1\}$ denoting if a data point $S_i$ is coming from a test/pseudo-test set
% ~($x=0$, unlabeled example) 
or a training set.
% ~($x=1$, labeled example).
Our goal is to infer the class labels of the data points from a test set, $\bs{y}_*$.

% \vspace{-1mm}
\subsection{Hashcode Representations}
% \vspace{-1mm}
% 
As per the hashcode representation approach, $\bs{S}$ is mapped to a set of \emph{locality sensitive} hashcodes, $\bs{C} = \{ \bs{c}_i\}_1^N$, using a set of $H$ binary hash functions, i.e. $\bs{c}_i = \bs{h}(S_i) = \{ h_1(S_i), \cdots, h_H(S_i) \}$.  
% 
% These hashcodes serve as feature vector representations so as to be used with any classification model, for instance, a random forest of decision trees. 
% In the following, we describe the concept of locality sensitivity, discuss a general procedure to learn a set of locality sensitive hash functions.
% 
% \subsubsection{Locality sensitivity of hash functions}
% 
% With locality sensitive hash functions, the probability of two examples to be assigned same hashcodes is proportional to their similarity~(defined as per any valid similarity/distance function); this also implies that examples that are very similar to each other are assigned hashcodes with minimal Hamming distance to each other, and vica versa~\cite{wang2014hashing,wang2017survey,indyk1998approximate}. Locality sensitivity of hash functions serve as the basis for using hashcodes as representations, for nearest neighbor search~\cite{garg2019kernelized,garg2018stochastic,kulis2012kernelized,joly2011random}, or even for clustering as we propose in \secref{sec:ssh}.
% 
% , we have an additional use of locality sensitive hash functions for clustering of examples to facilitate semi-supervised representation learning.
% 
% \subsubsection{Basics of constructing hash functions}
% 
% A locality sensitive hash function, 
$h_l(.; \bs{\theta})$ is constructed such that it splits a set of data points, $\bs{S}^R_l$, into two subsets as shown in \figref{fig:hash_choices}, while choosing the set as a small random subset of size $\alpha$ from the superset $\bs{S}$, i.e. $\bs{S}^R_l \subset \bs{S}$ s.t. $|\bs{S}^R_l|=\alpha \ll N$. In this manner, we can construct a large number of hash functions, $\{ h_l(.; \bs{\theta}) \}_{l=1}^H$, from a \emph{reference set} of size $M$, $\bs{S}^R = \{ \bs{S}^R_1 \cup \cdots \cup \bs{S}^R_H \}, |\bs{S}^R| = M \leq \alpha H \ll N$. 
% Although a hash function is constructed from a handful of examples, the parameters, $\bs{\theta}$, and optionally, the selection of $\bs{S}^R$ can be optimized using the whole training dataset, $\{ \bs{S}_T, \bs{y}_T \}$.

% \subsubsection{Supervised learning of kernelized hash functions}
% 
While, mathematically, a locality sensitive hash function can be of any form, kernelized hash functions~\cite{garg2019kernelized,garg2018stochastic,joly2011random},
% 
% previously introduced locality sensitive hash functions, which are applicable for hashing of natural language structures, are based on kernel methods~\cite{garg2019kernelized,garg2018stochastic,joly2011random}, 
rely upon a convolution kernel similarity function $K(S_i,S_j; \bs{\theta})$ defined for any pair of structures $S_i$ and $S_j$ with kernel-parameters $\bs{\theta}$~\cite{haussler1999convolution}. To construct $h_l(.)$, a kernel-trick based model, such as kNN, SVM, is fit to $\{ \bs{S}^R_l, \bs{z}_l \}$, with a randomly sampled binary vector, $\bs{z}_l \in \{0, 1\}^{\alpha}$, that defines the split of $\bs{S}^R_l$.
For computing hashcode $\bs{c}_i$ for $S_i$, it requires only $M$ number of convolution-kernel similarities of $S_i$ w.r.t. the data points in $\bs{S}^R$, which makes this approach highly scalable in compute cost terms.
% % 
% \vspace{-3mm}
% \begin{align}
% % \bs{\theta}^*, {\bs{S}^R}^*
% % \gets
% \mathop{\arg\max}
% \limits_{\bs{\theta},\ \bs{S}^R: \bs{S}^R \subset \bs{S}_T}
% \cI(\bs{c}: y);\ 
% \bs{c} = \bs{h}(S; \bs{\theta}, \bs{S}^R)
% \label{eqn:opt_global_par_supervised}
% \vspace{-3mm}
% \end{align}
    
In the previous work~\cite{garg2019kernelized}, it is proposed to optimize all the hash functions jointly by learning only the \emph{parameters which are shared amongst all the functions}, i.e. learning kernel parameters, $\bs{\theta}$ and the choice of reference set, $\bs{S}^R \subset \bs{S}_T$. This optimization is performed in a supervised manner via maximization of the mutual information between hashcodes of data points and their class labels, using $\{  \bs{S}_T, \bs{y}_T \}$ for training.
% ~\eqnref{eqn:opt_global_par_supervised}

% 
% Further, although it is difficult to characterize mathematically, the inherent randomness in sampling of a subset is constitutive towards preserving locality sensitivity of hash functions. In the light of this important algorithmic nature of the core approach, in our proposed framework, sampling of a subset is pseudo-random, i.e. random sampling of examples locally from within one of the clusters of the examples in the superset.

% \vspace{-1mm}
\section{Nearly-Unsupervised Hashcode Representations}
\label{sec:ssh}
% \vspace{-1mm}
    
\begin{figure*}[tp!]
\centering
\subfigure[
% \csize
Global sampling
]{
\includegraphics[
width=0.64\columnwidth]{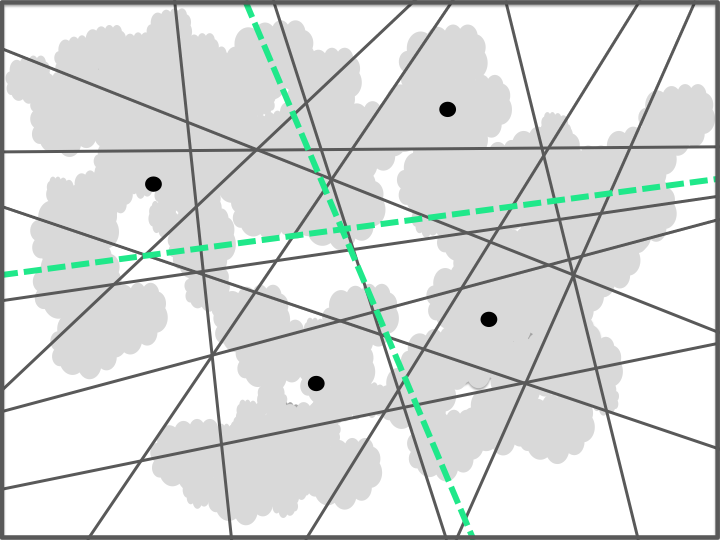}
\label{fig:hash_global}
}
%
% \hspace{5mm}
% 
\subfigure[
% \csize
Local sampling from a cluster
]{
\includegraphics[
width=0.64\columnwidth]{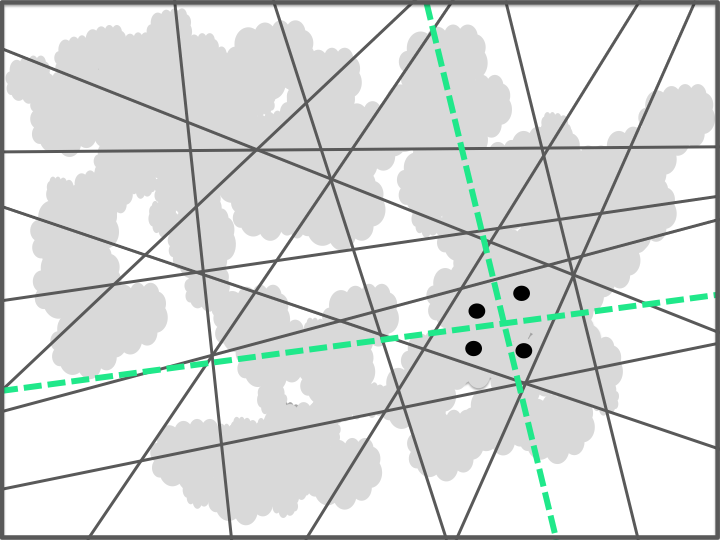}
\label{fig:hash_local}
}
% 
% \hspace{5mm}
% 
\subfigure[
% \csize
Local sampling from a high-entropy cluster]{
\includegraphics[
width=0.64\columnwidth]{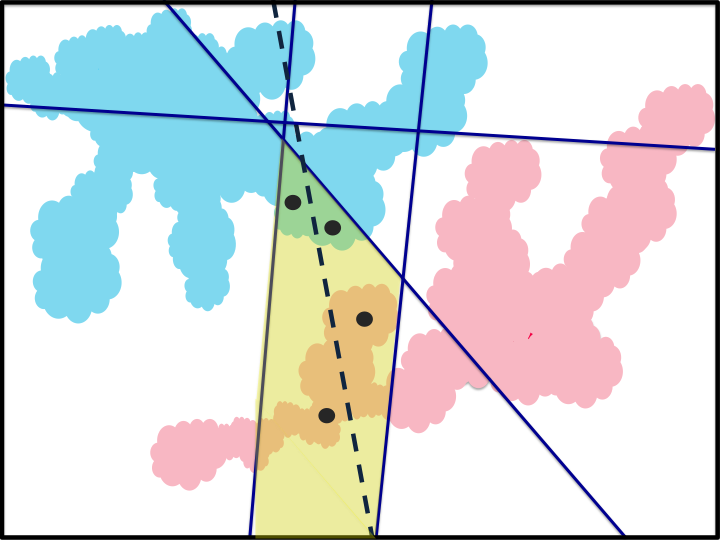}
\label{fig:high_entropy}
}
%
% \vspace{-4.75mm}
%
\caption{
% \csizeten
% 
% We illustrate the idea of sampling examples locally from within a cluster which are used to construct a hash function. 
% 
In all the three figures, dark-gray lines denote hash functions optimized previously, and intersections of the lines give us 2-D cells denoting hashcodes as well as clusters; black dots represent a set of data points which are used to construct a hash function, and some of the choices to split the subset are shown as green dashed lines. The procedure of sampling the subset varies across the three figures.
% 
% In \figref{fig:hash_global} and \figref{fig:hash_local}, gray color represent training or test examples, and  The intersections of the lines give us 2-D cells, denoting hashcodes as well as clusters. In \figref{fig:hash_global}, examples to construct a hash function are sampled globally, whereas in \figref{fig:hash_local}, examples are sampled locally from within a cluster, and some of the choices to split the subset are shown with green dashed lines. 
% 
In \figref{fig:hash_global}, since the four data points from global sampling have already unique hashcodes~(2-D cells), the newly constructed hash function~(one of the green dashed-lines) adds little information to their representations. On the other hand in \figref{fig:hash_local}, a hash function constructed from a set of data points, which are sampled locally from within a cluster, puts some of the data points in the sampled set into two different cells~(hashcodes), so adding more fine-grained information to their representations, hence more advantageous from representation learning perspective.
% 
% So it can be advantageous to construct some of the hash functions from local sampling. Further local sampling can be refined with the knowledge of which examples belongs to a training set vs a test set, as shown in \figref{fig:high_entropy}.
% 
In \figref{fig:high_entropy}, training and test data points are denoted with blue and red colors respectively, and the data points to construct a new hash function are sampled 
% The solid lines are the hash functions optimized previously. In a greedy step of our information-theoretic algorithm, a hash function is constructed by optimally splitting a very small subset of examples into two parts. 
% In the figure, 
% four examples are sampling randomly from with a cluster~(shaded 2-D cell), and the hash function is shown with a thick black-dashed line. We suggest that that it is a safer choice to sample 
locally from within a high entropy cluster, i.e. the one containing a balanced proportion of the training \& test data points.}
% 
% \vspace{-5.25mm}
% 
\label{fig:klsh_ssh_hierarchical}
\end{figure*}

% \paragraph{Limitations of the supervised hashcode learning approach}
% 
Our key insight in regards to limitation of the previous approach for supervised learning of hashcode representations, 
% also mentioned in their paper~\cite{garg2019kernelized}, 
is that, to avoid overfitting, learning is intentionally restricted only to the optimization of shared parameters whereas each hash function $h_l(.)$ is constructed in a randomized manner, i.e. random sub-sampling of a subset, $\bs{S}^R_l \subset \bs{S}$, and a random split of the subset. 
On the other hand, in a nearly-unsupervised hashcode learning settings as we introduce next, we can have the additional knowledge of data points from a test/pseudo-test set which can be leveraged to extend the optimization from the shared (global) parameters to fine-grained optimization of hash functions, not only to avoid overfitting but for higher generalization of hashcodes across training \& test sets.
% 
% due to the knowledge of data points from a test/pseudo-test set.
% , as we demonstrate in \secref{sec:ssh}.
    % 
% In this section, we propose a novel algorithm for a fine-grained optimization of the hash functions within our semi-supervised~(nearly unsupervised) learning framework, with the objective of learning representations generalizing across training \& test sets.
    
\paragraph{Nearly unsupervised learning settings\\}
We propose to learn hash functions, $\bs{h}(.)$, using $\bs{S} = \{\bs{S}_T, \bs{S}_*\}$, $\bs{x}$, and optionally, $\bs{y}_T$. Herein, $\bs{S}_*$ is a test set, or a pseudo-test set that can be a random subset of the training set or a large set of unlabeled data points outside the training set. 
% In the latter case, representation learning becomes purely unsupervised.
% 
% One can optionally use the class labels from the training set, $\bs{y}_T$, which corresponds to semi-supervised learning settings, transductive and inductive respectively.

% of data points for which we want to infer class labels $\bs{y}_*$, corresponding to transductive (semi-supervised) settings. One can also consider inductive  (semi-supervised) settings where $\bs{S}_*$ is a~(random) subset of a training set, or it is a very large set of unlabeled data points outside a training set. Within the semi-supervised setting, inductive or transductive, if we are not using $\bs{y}_T$ for learning $\bs{h}(.)$, but only, $\bs{S}, \bs{x}$, we refer to it as \emph{nearly unsupervised} settings. 

% In the next section, we introduce our framework for learning hash functions in above specified learning settings, to optimize global parameters as well optimizing aspects local to a hash function. After learning hash functions, one can compute hashcodes $\bs{C}$ for $\bs{S}$, and train a random forest classifier, or any other classifier, using $\{ \bs{C}_T, \bs{y}_T \}$ to infer class labels for input of hashcodes, $\bs{C}_*$, computed for $\bs{S}_*$. One can also train a semi-supervised classifier, using $\{ \bs{C}, \bs{x}, \bs{y}_T \}$.

% \vspace{-1mm}
\subsection{Basic Concepts for Fine-Grained Optimization}
% \vspace{-1mm}
% 
\label{sec:basic_concepts_ssh}

In the prior works on kernel-similarity based locality sensitive hashing, the first step for constructing a hash function is to randomly sample a small subset of data points, from a superset $\bs{S}$, and in the second step, the subset is split into two parts using a \emph{kernel-trick} based model~\cite{garg2019kernelized,garg2018stochastic,joly2011random}, serving as the hash function, as described in \secref{sec:ps}.
% 
% , or it is used to find an approximately linear-hyperplane in the kernel implied feature space~\cite{kulis2009kernelized,kulis2012kernelized}. In our approach, we focus on the former technique of factorizing a subset into two parts for constructing a hash functions, since it is more generic, corresponding to a variety of hash functions, such as Random Maximum Margin, Random K Nearest Neighbors, etc. 
    
In the following, we introduce basic concepts for improving upon these two key aspects of constructing a hash function, while later, in \secref{sec:semisupervised_hash_opt}, these concepts are incorporated in a unified manner in our proposed \emph{information-theoretic algorithm that greedily optimizes hash functions} one by one.
    
% \vspace{-1mm}
\paragraph{Informative split\\}
In \figref{fig:hash_choices}, construction of a hash function is pictorially illustrated, showing multiple possible splits, as dotted lines, of a small set of four data points~(black dots).~(Note that a hash function is shown to be a linear hyperplane only for simplistic explanations of the basic concepts.) While in the previous works, one of the many choices for splitting the set is chosen randomly, we propose to optimize upon this choice. Intuitively, one should choose a split of the set, corresponding to a hash function, such that it gives a balanced split for the whole set of data points, and it should also generalize across training \& test sets. In reference to the figure, one simple way to analyze the generalization of a split~(so the hash function) is to see if there are training as well as test data points~(or pseudo-test data points) on either side of the dotted line. As per this concept, an optimal split of the set of four data points is shown in \figref{fig:hash_choice_optimal}.
    
Referring back to \secref{sec:ps}, clearly, this is a combinatorial optimization problem, where we need to choose an optimal choice of $\bs{z}_l \in \{0, 1\}^{\alpha}$ for set, $\bs{S}^R_l$, to construct $h_l(.)$. For a small value of $\alpha$, one can either go through all the possible combinations in a brute force manner, or use Markov Chain Monte Carlo sampling.
It is interesting to note that, even though a hash function is constructed from a very small set of data points~(of size $\alpha \not \gg 1$), the generalization criterion, formulated in our info-theoretic objective introduced in \secref{sec:semisupervised_hash_opt}, is computed using all the data points available for the optimization, $\bs{S}$.
    
% \vspace{-1mm}
\paragraph{Local sampling from a cluster\\}
Another aspect of constructing a hash function, having a scope for improvement, is sampling of a small subset of data points, $\bs{S}^R_l \subset \bs{S}$, that is used to construct a hash function. In the prior works, the selection of such a subset is purely random, i.e. random selection of data points globally from $\bs{S}$.     
In \figref{fig:klsh_ssh_hierarchical}, we illustrate that, 
% after constructing a small number of hash functions with purely random sub-sampling of the subsets, 
it is wiser to (randomly) \emph{select data points locally from one of the clusters} of the data points in $\bs{S}$, rather than sampling globally from $\bs{S}$. Here, we propose that clustering of all the data points in $\bs{S}$ can be obtained using the hash functions itself, due to their locality sensitive property. 
While using a large number of locality sensitive hash functions give us fine-grained representations of data points, \emph{a small subset of the hash functions, of size $\zeta$, defines a valid clustering} of the data points, since data points which are similar to each other should have same \emph{hashcodes serving as cluster labels}. 
    
From this perspective, we can construct first few hash functions from global sampling of data points, what we refer as global hash functions. These global hash functions should serve to provide hashcode representations as well as clusters of data points. Then, via local sampling from the clusters, we can also construct \emph{local hash functions} to capture more finer details of data points.
% , which may be missed out by global hash functions. 
    % 
As per this concept, we can learn \emph{hierarchical (multi-scale) hashcode representations} of data points, capturing differences between data points from coarser~(global hash functions) to finer scales~(local hash functions). 
    
Further, we suggest to choose a cluster
% two criteria to choose a cluster for local sampling: (i) a cluster should have a reasonably high number of data points belonging to it; (ii) 
that has a balanced proportion of training \& test~(pseudo-test) data points, 
% so that a new cluster emerging from the split has a high probability of containing b, 
which is desirable from the perspective of having generalized hashcode representations; see
% \figref{fig:not_agnostic}, \ref{fig:agnostic}, 
\figref{fig:high_entropy}.
            
% \vspace{-1mm}
\paragraph{Splitting other clusters\\}
In reference to \figref{fig:factor_clusters}, \emph{non-redundancy} of a hash function w.r.t. the other hash functions can be characterized in terms of how well the hash function splits the clusters defined as per the other hash functions.

\paragraph*{}
Next we mathematically formalize all the concepts introduced above for fine-grained optimization of hash functions into an information-theoretic objective function.

% \vspace{-1mm}
\subsection{Information-Theoretic Learning}
\label{sec:semisupervised_hash_opt}
% \vspace{-1mm}

\begin{algorithm*}[tp!]
\caption{
% \csizenine
Nearly-Unsupervised Hashcode Representation Learning
}
% 
% \csizeten
% 
\begin{algorithmic}[1]
\REQUIRE Dataset $\{ \bs{S}, \bs{x} \}$, and parameters, $H, \bs{\alpha}, \zeta$.
\STATE $\bs{h}(.) \gets \{ \}$, 
$\bs{C} \gets \{\}$, $\bs{f} \gets \{\}$\\
% , $\bs{S}^R \gets \{ \}$\\
% \\
% 
\COMMENT{Greedy step for optimizing hash function, $h_l(.)$}
% 
% \FOR{$l=1 \to H$}
% 
\WHILE{$|\bs{h}(.)| < H$}
\STATE $\alpha_l \gets$ sampleSubsetSize$\left( \bs{\alpha} \right)$
\COMMENT{Sample a size for the subset of data points, $\bs{S}^R_l$, to construct a hash function}
\IF{$|\bs{h}(.)| < \zeta$}
\STATE $\bs{S}^R_l \gets$ randomSampleGlobally$\left( \bs{S}, \alpha_l \right)$
\COMMENT{Randomly sample data points, globally from $\bs{S}$, for constructing a global hash function.}
\ELSE
\algshade{
\STATE $\bs{S}^R_l \gets$ randomSampleLocally$\left( \bs{S}, \bs{C}, \bs{x}, \alpha_l, \zeta \right)$
\COMMENT{Sample data points randomly from, a high entropy cluster, for constructing a local hash function.}
}
\ENDIF
\\
\algshade{
\STATE $h_l(.), f_l \!\gets\!$ optimizeSplit$\left( \bs{S}^R_l, \bs{S}, \bs{C}, \bs{x} \right)$
\COMMENT{Optimize split of $\bs{S}^R_l$.}
}
\STATE $\bs{c} \gets$ computeHash$\left( \bs{S}, h_l(.) \right)$
\STATE $\bs{C} \gets \bs{C} \cup \bs{c}$, 
$\bs{h}(.) \gets \bs{h(.)} \cup h_l(.)$, 
$\bs{f} \gets \bs{f} \cup f_l$.
\STATE $\bs{h}(.), \bs{C}, \bs{f} \gets$ deleteLowInfoFunc$\left(\bs{h}(.), \bs{C}, \bs{f} \right)$
\COMMENT{delete hash functions from the set with lower objective 
% function 
values
}
\ENDWHILE
\STATE \textbf{Return} $\bs{h}(.), \bs{C}$.
\end{algorithmic}
\label{alg:opt_ref}
\end{algorithm*}

\begin{figure}[tp!]
\centering
% 
% \subfigure[Global sampling]{
% \includegraphics[
% width=0.48\columnwidth]{global_hash.png}
% \label{fig:hash_global}
% }
%
% \hspace{-2mm}
% 
% \subfigure[Local sampling]{
\includegraphics[
width=0.8\columnwidth]{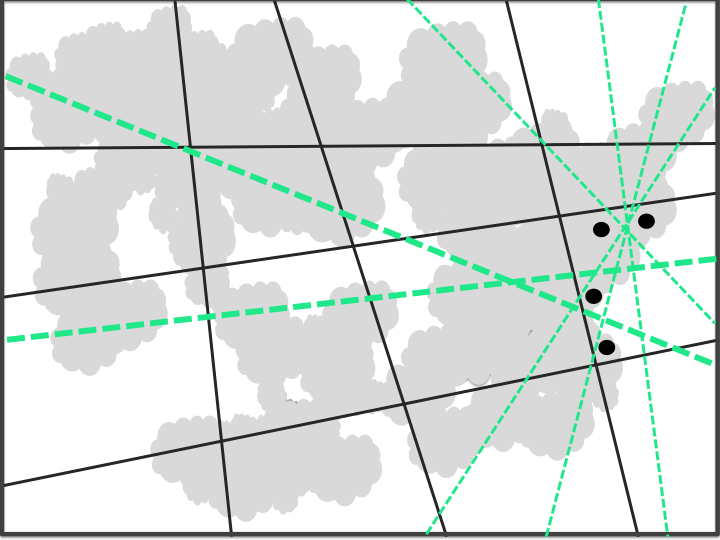}
% }
% 
% \vspace{-2.0mm}
% 
\caption{
% \csizeten
% 
Dark-gray lines denote hash functions optimized previously,
% in a greedy algorithmic setting, 
and gray color represents data points used for the optimization. The intersections of the lines give us 2-D cells, corresponding to hashcodes as well as clusters. For the set of four data points sampled within a cluster, there are many choices to split it~(corresponding to hash functions choices), shown with dashed lines. We propose to choose the one which also splits the sets of data points in other (neighboring) clusters in a balanced manner, i.e. having data points in a cluster on either side of the dashed line and cutting through as many clusters as possible. As per this criterion, the thick-green dashed lines are superior choices w.r.t. the thin-green ones.
% 
% a measure to capture disentanglement~(redundancy) of the hash function~(the one being optimized) w.r.t. the other hash functions~(optimized previously).
}
% \vspace{-4mm}
\label{fig:factor_clusters}
\end{figure}

We optimize hash functions greedily, one by one. Referring back to \secref{sec:ps}, we define binary random variable $x$ denoting if a data point comes from a training set or a test/pseudo-test set. In a greedy step of optimizing a hash function, random variable, $\bs{c}$, represents the hashcode of a data point $S$, as per the previously optimized hash functions $\bs{h}_{l-1}(.)$.
% 
% = \{ h_1(.), \cdots, h_{l-1}(.) \}$; $\bs{c} = \bs{h}_{l-1}(S)$. 
% 
Along same lines, $c$ denotes the binary random variable corresponding from the present hash function under optimization, $h_l(.)$.
% ; $c = h_l(S)$. 
% 
We maximize the information-theoretic objective as below.

% \vspace{-1mm}
\begin{align}
% &
% \arg\!\max_{h_l()} H(x,c) + H(x, \bs{c}) + H(c, \bs{c});
% \ \bs{c} = \bs{h}_{l-1}(S),
%  c = h_{l-1}(S)
% \\
% &
% &
\arg\!\max_{h_l()} \cH(x,c) -\cI(c:\bs{c}) + \cH(x|\bs{c});
\\
\nonumber
% &
\bs{c} = \bs{h}_{l-1}(S),
 c = h_{l}(S)
\end{align}

\noindent Herein the optimization of a hash function, $h_(.)$, involves intelligent selection of $\bs{S}^R_l$, an informative split of $\bs{S}^R_l$, i.e. optimizing $\bs{z}_l$ for $\bs{S}^R_l$, and learning of the parameters $\bs{\theta}$ of a (kernel or neural) model, which is fit on $\{ \bs{S}^R_l, \bs{z}_l \}$, acting as the hash function.
            
In the objective function above, maximizing the first term, $\cH(x, c)$, i.e. joint entropy on $x$ and $c$,
% means that the hash function should be highly randomized, having a balanced split of the examples which are used for optimization. The second term, $\cH(x|c)$, 
corresponds to the concept of \emph{informative split} described above in \secref{sec:basic_concepts_ssh};
% , i.e. a hash function assigning value 0 to labeled~(training) as well as unlabeled examples~(test), and same applies for value 1; 
see \figref{fig:klsh_ssh_concepts}. This term is cheap to compute since $x$ and $c$ are both 1-dimensional binary variables.
    
The second term in the objective, the mutual information term, ensures minimal redundancies between hash functions.
% corresponding to the concept of splitting clusters.
% 
This is related to the concept of constructing a hash function
% from examples sampled within a cluster, 
such that it splits many of the existing clusters, as mentioned above in \secref{sec:basic_concepts_ssh}; see \figref{fig:factor_clusters}.
This mutual information function can be computed using the approximation in the previous work by ~\cite{garg2019kernelized}.
        
The last quantity in the objective is $\cH(x|\bs{c})$, conditional entropy on $x$ given $\bs{c}$. 
% Since both quantities, $x$ and $\bs{c}$ are fixed in the optimization, this term is not directly effected from the optimization of hash function, $h_l(.)$, as such. 
% 
% Yet, this term has a high significance from the perspective of learning generalized hashcode representations, since maximizing $\cH(x|\bs{c})$ encourages clusters, defined from the hash functions, to contain labeled~(training) as well as unlabeled~(test) examples. 
% 
% So 
We propose to maximize this term indirectly via \emph{choosing a cluster informatively, from which to randomly select data points for constructing the hash function}, such that it contains a balanced ratio of the count of training \& test data points, i.e. a cluster with high entropy on $x$, which we refer as a \emph{high entropy cluster}.
        % 
% As explained in \secref{sec:basic_concepts_ssh}, 
% The hash function constructed from data points within a cluster 
% splits the clusters itself, and 
% 
In reference to \figref{fig:high_entropy}, the new clusters emerging from a split of a high entropy cluster should have higher chances to be high entropy clusters themselves, thus maximizing the last term indirectly.
% in comparison to the ones emerging from a split of a lower entropy cluster. 
% See \figref{fig:high_entropy} to understand this concept pictorially.
% Computations in the procedure of characterizing entropies of clusters is cheap, requiring to 
We compute marginal entropy on $x$ for each cluster, and an explicit computation of $H(x|\bs{c})$ is not required.
    
% \paragraph{Nearly Unsupervised}
% 
% It is interesting to observe that the above info-theoretic objective uses only the knowledge of whether an example is labeled or unlabeled~(random variable $x$), while ignoring the actual class label of an example. In this sense, our approach is \emph{nearly unsupervised}. Our emphasis is on fine-grained optimization of each hash function such that hashcode representations generalize across training \& test sets. 
% 
Optionally one may extend the objective to include the term, $-\cH(y|\bs{c}, c)$, with $y$ denoting the random variable for a class label.
    
% \vspace{-2mm}
% \begin{align}
% \arg\!\max_{h_l()} \cH(c) \!+\! \cH(x|c) \!+\! \cH(c|\bs{c}) \!+\! \cH(x|\bs{c}) \!-\! \cH(y|\bs{c}, c);
% \\
% \nonumber
% % &
% \bs{c} = \bs{h}_{l-1}(S),
%  c = h_{l}(S)
% \end{align}

% \subsubsection{Algorithmic Details}

The above described learning framework is summarized in \algoref{alg:opt_ref}. 
    
% \paragraph{Neural network as a hash function\\}
% 
% For optimizing a kernelized hash function given $\bs{S}^R_l$, $h_l(.)$, when finding an optimal choice of $\bs{z}_l \in \{0, 1\}^{\alpha}$ for $\bs{S}^R_l$, we also optimize kernel parameters, $\bs{\theta}$, of a kernel-trick based model, which is fit on $\bs{S}^R_l$ along with a given choice of $\bs{z}_l$, corresponding to one of the choices for $h_l(.)$. Thus we learn kernel parameters local to a hash function unlike the previous works. 
% Further, 
% 
Besides kernelized locality sensitive hashcodes, the above framework allows neural locality sensitive hashing. One can fit any ~(regularized) neural model on $\{ \bs{S}^R_l, \bs{z}_l \}$, acting as a \emph{neural locality sensitive hash function}. 
% For a random choice of $\bs{z}_l$ for $\bs{S}^R_l$, one can expect that the neural hash function may overfit to $\{ \bs{S}^R_l, \bs{z}_l \}$, so acting as a poor hash function. However it is reasonable to 
We expect that some of the many possible choices for a split of $\bs{S}^R_l$ should lead to natural semantic categorizations of the data points. For such a natural split choice, even a parameterized model can act as a good hash function without overfitting  as we observed empirically.
% 
% . The proposed objective function accomplishes the goal of 
% , leading to obtain even with a highly parameterized neural model, as we observe empirically.

% \paragraph{Deleting (weak) hash functions\\}
% 
In the algorithm, we also propose to \emph{delete some of the hash functions} from the set of optimized ones, the ones which have low objective function values w.r.t. the rest. This step provides robustness against an arbitrarily bad choice of \emph{randomly} selected subset, $\bs{S}^R_l$.

Our algorithm allows parallel computing, as in the previous hashcode learning approach.

\section{Experiments}
% \vspace{-1mm}

We demonstrate the applicability of our approach for a challenging task of biomedical relation extraction, using four public datasets.
    
% \vspace{-1mm}
\paragraph{Dataset details\\}
For AIMed and BioInfer, cross-corpus evaluations have been performed in many previous works~\cite{airola2008all,tikk2010comprehensive,peng2017deep,hsieh2017identifying,rios2018generalizing,garg2019kernelized}. These datasets have annotations on pairs of interacting proteins~(PPI) in a sentence while ignoring the interaction type. Following the previous works, for a given pair of proteins mentioned in a text sentence from a training or a test set, we obtain the corresponding undirected shortest path from a Stanford dependency parse of the sentence, that serves as a data point.
        
We also use PubMed45 and BioNLP datasets which have been used for extensive evaluations in recent works~\cite{garg2019kernelized,garg2018stochastic,raobiomedical,garg2016extracting}. These two datasets consider a relatively more difficult task of inferring an interaction between two or more bio-entities mentioned in a sentence, along with the inference of their interaction-roles, and the type of interaction from an unrestricted list. As in the previous works, we use abstract meaning representation~(AMR) to obtain shortest path-based data points~\cite{Banarescu13abstractmeaning}; same bio-AMR parser~\cite{pust2015parsing} is employed as in the previous works. 
PubMed45 dataset has 11 subsets, with evaluation performed for each of the subsets as a test set leaving the rest for training~(not to be confused with cross-validation). For BioNLP dataset~\cite{kim2009overview,kim2011overview,nedellec2013overview}, the training set contains annotations from years 2009, 2011, 2013, and the test set contains development set from year 2013. Overall, for a fair comparison of the models, we keep same experimental setup as followed in \cite{garg2019kernelized},
for all the four datasets, so as to avoid any bias due to engineering aspects; evaluation metrics for the relation extraction task are, f1 score, precision, recall.
    
% \vspace{-1mm}
\paragraph{Baseline methods\\}
The most important baseline method for the comparison is the recent work of supervised hashcode representations~\cite{garg2019kernelized}. Their model is called as \emph{KLSH-RF}, with KLSH referring to kernelized locality sensitive hashcodes, and RF denotes Random Forest. Our approach differs from their work in the sense that our hashcode representations are nearly unsupervised, whereas their approach is purely supervised, while both approaches use a supervised RF. We refer to our model as \emph{KLSH-NU-RF}. Within the nearly unsupervised learning setting, we consider transductive setting by default, i.e. using data points from both training and test sets. Later, we also show results for inductive settings, i.e. using a random subset of training data points, as a pseudo-test set. In both scenarios, we do not use class labels for learning hashcodes, but only for training RF. 

\begin{table}[tp!]
\centering
% \tabsize
\tabsize
\renewcommand{\tabcolsep}{3.6pt}
\begin{tabular}{lll}
\toprule
\textbf{Models}
&\textbf{(A, B)}&\textbf{(B, A)}\\
\toprule
% % % 
% SVM$_1$~(Airola08)&0.25&0.44\\
% \midrule
% % 
% SVM$_2$~(Airola08)&0.47&0.47\\
% \midrule
% 
SVM~(Airola08)&0.47&0.47\\
\midrule
% %
SVM~(Miwa09)&0.53&0.50\\
\midrule
% 
% % 
% RF&0.30&0.52\\
% &(0.64, 0.19)&(0.47, 0.59)\\
% \toprule
% 
% 
SVM~(Tikk10)
% ~\cite{tikk2010comprehensive}
&0.41&0.42\\
&(0.67, 0.29)&(0.27, 0.87)\\
\toprule
% 
% \midrule
CNN~(Nguyen15)
% ~\cite{nguyen2015relation}
&0.37&0.45\\
\midrule
Bi-LSTM~(Kavuluru17)
% ~\cite{kavuluru2017extracting}
&0.30&0.47\\
\midrule
CNN~(Peng17)
% ~\cite{peng2017deep}
&0.48&0.50\\
&(0.40, 0.61)&(0.40, 0.66)\\
\midrule
RNN~(Hsieh17)
% ~\cite{hsieh2017identifying}
&0.49&0.51\\
\toprule
CNN-RevGrad~(Ganin16)
% ~\cite{ganin2016domain}
% ~\cite{rios2018generalizing}
&0.43&0.47\\
\midrule
Bi-LSTM-RevGrad~(Ganin16)
% ~\cite{ganin2016domain}
&0.40&0.46\\
\midrule
Adv-CNN~(Rios18)
% ~\cite{rios2018generalizing}
&0.54&0.49\\
\midrule
Adv-Bi-LSTM~(Rios18)
% ~\cite{rios2018generalizing}
&$\bs{0.57}$&0.49\\
\toprule
KLSH-kNN~(Garg18)
% ~\cite{garg2018stochastic}
&$0.51$&$0.51$\\
&(0.41, 0.68)&(0.38, 0.80)\\
\midrule
KLSH-RF~(Garg19)
% ~\cite{garg2019kernelized}
&$\bs{0.57}$&$0.54$\\
&(0.46, 0.75)&(0.37, 0.95)\\
\midrule
SSL-VAE~(Zhang19)&0.50&0.46\\
&(0.38, 0.72)&(0.39, 0.57)\\
\toprule
% 
% B,A
% RMM, alpha=3
% R=100
% H=100
% exact 0.5 threshold, no need to tune, unlike prior work
% 0.567 (0.43, 0.82)
%
% RkNN, alpha=4
% 0.559 (0.44, 0.81)
% 
\textbf{KLSH-NU-RF}
&$\bs{0.57}$&$\bs{0.57}$\\
&(0.44, 0.81)&(0.44, 0.81)\\
% 
% \midrule
% % 
% NLSH-NU-RF
% &$0.56$&$0.55$\\
% &(0.43, 0.80)&(0.40, 0.84)\\
% 
\toprule
\end{tabular}
% \vspace{-3mm}
\caption{
% 
% \csizeten
% 
% We show 
Evaluation results from cross-corpus evaluation for (train, test) pairs of datasets, AIMed~(A) and BioInfer~(B). For each model, we report F1 score, and
% in the first row, and corresponding to it. 
% In some of the previous works, precision, recall numbers are not reported; 
if available, precision, recall scores are also shown in brackets. For the adversarial neural models by \cite{ganin2016domain}, evaluation on the datasets was provided by \cite{rios2018generalizing}.
% 
% Here, ``Ganin16-Rios18" means that the model is originally proposed in \protect\cite{ganin2016domain}, and evaluated for these datasets by \protect\cite{rios2018generalizing}.
}
% \vspace{-3mm}
\label{tab:results_ppi}
\end{table}

For AIMed and BioInfer datasets, adversarial learning based four neural network models had been evaluated in the prior works~\cite{rios2018generalizing,ganin2016domain}, referred as \emph{CNN-RevGrad, Bi-LSTM-RevGrad, Adv-CNN, Adv-CNN}. Like our model KLSH-NU-RF, these four models are also learned in transductive settings of using data points from the test set in addition to the training set. Semi-supervised Variational Autoencoders~(\emph{SSL-VAE}) have also been explored for biomedical relation extraction~\cite{zhang2019exploring}, which we evaluate ourselves for all the four datasets considered in this paper. 
% We also compare w.r.t. other supervised models, which had been evaluated previously on these datasets.
                     
% For our proposed approach of nearly unsupervised hashcode representations, we use a random forest~(RF) as the final supervised-classifier, following the previous work of \cite{garg2019kernelized} in which supervised hashcode representations are fed as input feature vectors to an RF~(KLSH-RF). We refer to our model as KLSH-NU-RF. 
% 
% For evaluation, we considered the task of biomedical relation extraction using four public datasets, AIMed, BioInfer, PubMed45, BioNLP, as done for KLSH-RF in their work. While PubMed45, BioNLP datasets represent more challenging aspects of the task, AIMed, BioInfer datasets have relevance because cross corpus evaluations been performed on these two datasets, using kernel as well as neural models including the ones doing adversarial training of neural networks using the knowledge of data points in a test set. For learning hashcode representations in our model, KLSH-NU-RF, we use training as well as test data points, unless mentioned otherwise.

% \vspace{-1mm}
\paragraph{Parameter settings\\}
We use path kernels with word vectors \& kernel parameter settings as in the previous work~\cite{garg2019kernelized}. From a preliminary tuning, we set the number of hash functions, $H=100$, and the number of decision trees in a Random Forest classifier, $R=100$; these parameters are not sensitive, requiring minimal tuning. For any other parameters which may require fine-grained tuning, we use 10\% of training data points, selected randomly, for validation. Within kernel locality sensitive hashing, we choose between Random Maximum Margin and Random k-Nearest Neighbors techniques, and for neural locality sensitive hashing, we use a simple 2-layer LSTM model with 8 units per layer. In our nearly unsupervised learning framework, we use subsets of the hash functions, of size 10, to obtain clusters~($\zeta=10$).
We employ 8 cores on an i7 processor, with 32GB memory, for all the computations.
    	
\begin{table}[tp!]
\centering
% \tabsize
\csize
\renewcommand{\tabcolsep}{6.7pt}
\begin{tabular}{llll}
\toprule
\textbf{Models}&\textbf{PubMed45}
%   &\textbf{PubMed45-ERN}
&\textbf{BioNLP}\\
%         &(Precision, Recall)\\
\toprule
SVM~(Garg16)
% ~\cite{garg2016extracting}
&$0.45$$\pm$$0.25$
% &$0.33$$\pm$$0.16$
&$0.46$\\
&(0.58, 0.43)
% &(0.33, 0.45)
&(0.35, 0.67)\\
\toprule
LSTM~(Rao17)
% ~\cite{raobiomedical}
&N.A.
% &N.A.
&0.46\\
&
% &
&(0.51, 0.44)\\
\midrule
LSTM~(Garg19)
% ~\cite{garg2019kernelized}
&$0.30$$\pm$$0.21$
% &$0.29$$\pm$$0.14$
&0.59\\
&(0.38, 0.28)
% &(0.42, 0.33)
&(0.89, 0.44)\\
\midrule
Bi-LSTM~(Garg19)
% ~\cite{garg2019kernelized}
&$0.46$$\pm$$0.26$
% &$0.37$$\pm$$0.15$
&0.55\\
&(0.59, 0.43)
% &(0.45, 0.40)
&(0.92, 0.39)\\
\midrule
LSTM-CNN~(Garg19)
% ~\cite{garg2019kernelized}
&$0.50$$\pm$$0.27$
% &$0.31$$\pm$$0.17$
&0.60\\
&(0.55, 0.50)
% &(0.35, 0.40)
&(0.77, 0.49)\\
\midrule
CNN~(Garg19)
% ~\cite{garg2019kernelized}
&$0.51$$\pm$$0.28$
% &$0.33$$\pm$$0.18$
&0.60\\
&(0.46, 0.46)
% &(0.36, 0.32)
&(0.80, 0.48)\\
\midrule
KLSH-kNN~(Garg18)
% ~\cite{garg2018stochastic}
&$0.46$$\pm$$0.21$
% &$0.23$$\pm$$0.13$
&$0.60$\\
&(0.44, 0.53)
% &(0.23, 0.29)
&(0.63, 0.57)\\
\midrule
KLSH-RF~(Garg19)
% ~\cite{garg2019kernelized}
&$0.57$$\pm$$0.25$
% &$0.45$$\pm$$0.22$
&$0.63$\\
&(0.63, 0.55)
% &(0.51, 0.52)
&(0.78, 0.53)\\
% 
% 
% \midrule
% LSTM&$0.48\pm0.32$
% % &$0.29\pm0.14$
% &$0.63$\\
% &(0.54, 0.47)
% % &(0.42, 0.33)
% &(0.78, 0.53)\\
% % 
% % 
% \midrule
% LSTM-CNN&$0.50\pm0.27$
% % &$0.\pm0.$
% &$0.63$\\
% &(0.55, 0.50)
% % &(0., 0.)
% &(0.78, 0.54)\\
% 
% \midrule
% GRU-CNN&$0.50\pm0.27$
% % &$0.\pm0.$
% &$$\\
% &(0.55, 0.50)
% % &(0., 0.)
% &(, )\\
% 
% 
\midrule
SSL-VAE~(Zhang19)&$0.40\pm0.16$&$0.48$\\
&(0.33, 0.69)&(0.43, 0.56)\\
\toprule
\textbf{KLSH-NU-RF}
&$\bs{0.61}$$\pm$$\bs{0.23}$
&$\bs{0.67}$\\
&(\textbf{0.61, 0.62})
% &(\textbf{, })
&(\textbf{0.73, 0.61})\\
% 
% \midrule
% \textbf{KLSH-NU$\bs{^i}$-RF}&$\bs{0.}$$\pm$$\bs{0.}$
% &$0.66$\\
% &(0., 0.)\\
% % &(\textbf{, })
% &(0.70, 0.62)
% 
% 
% \midrule
% NLSH-NU-RF&$0.59$$\pm$$0.23$
% &$0.63$\\
% &(0.57, 0.64)
% % &(\textbf{, })
% &(0.71, 0.56)\\
% 
% 
\toprule
\end{tabular}
% \vspace{-3mm}
\caption{
% \csizeten
Evaluation results for PubMed45  and BioNLP datasets. We report F1 score~(mean $\pm$ standard deviation), and mean-precision \& mean-recall numbers in brackets. For BioNLP, we standard deviation numbers are not provided as there is one fixed test subset.
}
% \vspace{-3mm}
\label{tab:results_pubmed45_bionlp_all}
\end{table}

% \vspace{-1mm}
\subsection{Experimental Results}
% \vspace{-1mm}
% 
In summary, our model KLSH-NU-RF 
% that is learned by nearly unsupervised optimization of hashcode representations which are fed as feature vectors into a supervised random forest classifier,
significantly outperforms its purely supervised counterpart, KLSH-RF, and also semi-supervised neural network models.

% \paragraph{AIMed and BioInfer Datasets}
% 
In reference to \tabref{tab:results_ppi}, we first discuss results for the evaluation setting of using AIMed dataset as a test set, and BioInfer as a training set. We observe that our model, KLSH-NU-RF, obtains F1 score, 3 pts higher w.r.t. the most recent baseline, KLSH-RF. In comparison to the semi-supervised neural neural models, CNN-RevGrad, Bi-LSTM-RevGrad, Adv-CNN, Adv-Bi-LSTM, SSL-VAE, which use the knowledge of a test set just like our model, we gain 8-11 pts in F1 score. 
% In addition, we evaluate semi-supervised Variational Autoencoders~(VAEs) as a competitive baseline ourselves on these datasets, and despite extensive tuning, we find their performance consistently inferior to our model. 
% 
On the other hand, when evaluating on BioInfer dataset as a test set and AIMed as a training set, our model is in tie w.r.t. the adversarial neural model, Adv-Bi-LSTM, though outperforming the other three adversarial models and SSL-VAE, by large margins in F1 score. In comparison to KLSH-RF, we retain same F1 score, while gaining in recall by 6 pts at the cost of losing 2 pts in precision.
    
For PubMed and BioNLP datasets, there is no prior evaluation of adversarial models. Nevertheless, in \tabref{tab:results_pubmed45_bionlp_all}, we see that our model significantly outperforms SSL-VAE, and it also outperforms the most relevant baseline, KLSH-RF, gaining F1 score by 4 pts for both the datasets.\footnote{These two datasets have high importance to gauge practical relevance of a model for the task of biomedical relation extraction.} 
Note that, high standard deviations reported for PubMed45 dataset are due to high diversity across the 11 test sets, while the performance of our model for a given test set is highly stable~(improvements are statistically significant with p-value: 6e-3).
    
One general trend to observe is that the proposed nearly unsupervised learning approach leads to a significantly higher recall, at the cost of marginal drop in precision, w.r.t. its supervised baseline.
        
% We also evaluate NLSH-NU-RF, and find it to be performing a little worse than it kernel based counterpart, KLSH-NU-RF, though it is highly competitive w.r.t. all the other baselines.

\begin{figure}[tp!]
\centering
\includegraphics[
width=\columnwidth]{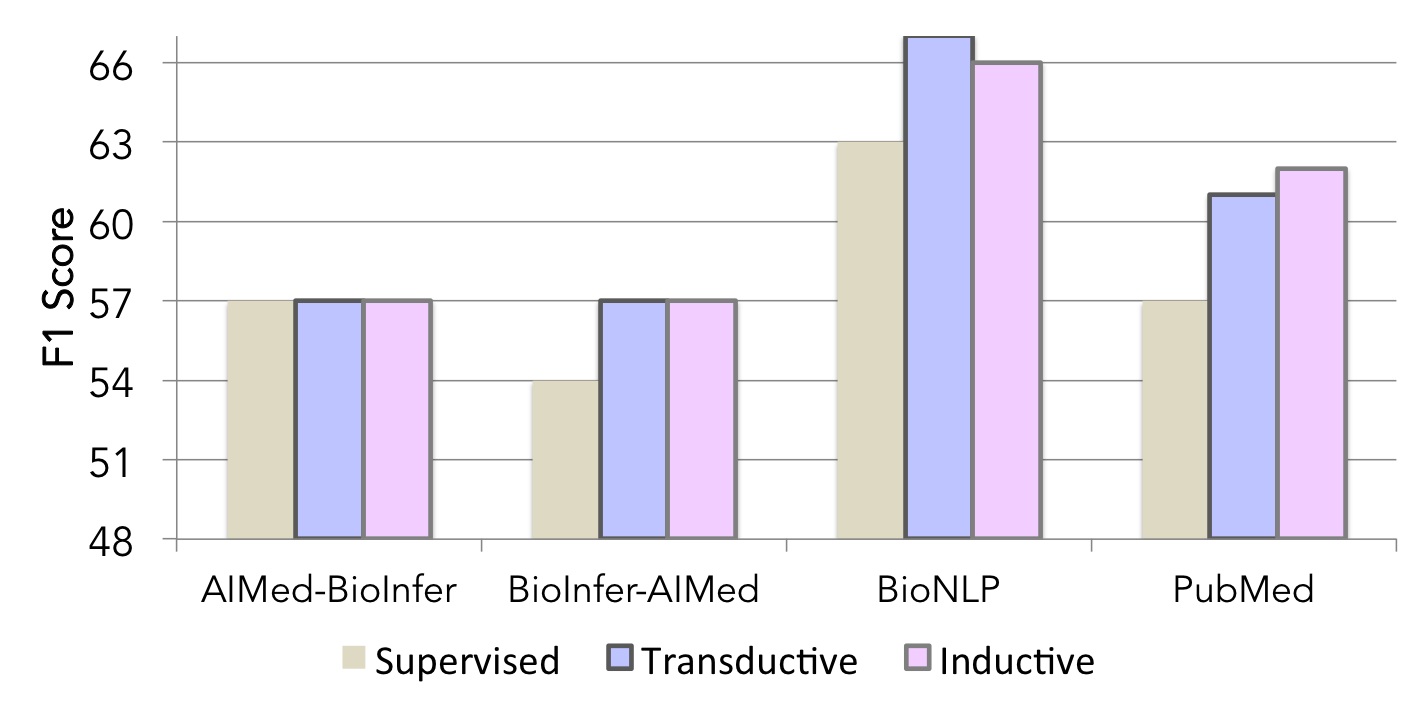}
% \vspace{-8mm}
\caption{
% \csizeten
Comparison across the three types of learning settings, supervised, transductive, and inductive.
}
% \vspace{-1mm}
\label{fig:hash_sit}
\end{figure}

% It is also noteworthy that, w.r.t. the most recent baseline for the four datasets used for the evaluation above, KLSH-RF, our model varies only in the sense that hashcode representations are learned in a nearly-unsupervised manner not only using data points from a training set but also a test set. As for building the final classifier, an RF, both approaches are purely supervised using only the hashcode representations of training examples along with their class labels, while completely ignoring the test set. Therefore it is quite interesting ~(and perhaps surprising) to observe such drastic improvements in F1 score w.r.t. the baseline, even more so considering the fact that in our nearly-unsupervised hashcode learning framework uses only the knowledge of which set a data points belongs to, a training or a test set, while completely ignoring the actual class labels from a training set. 

Further, note that the number of hash functions used in the prior work is 1000 whereas we use only 100 hash functions. Compute time is same as for their model.

Our approach is easily extensible for other modeling aspects such as non-stationary kernel functions, document level inference, joint use of semantic \& syntactic parses, ontology or database usage~\cite{garg2019kernelized,garg2016extracting,alicante2016distributed}, though we refrain from presenting system level evaluations, and have focused only upon analyzing improvements from our principled extension of the recently proposed technique that has already been shown to be successful for the task.

% \vspace{-1mm}
\paragraph{Transductive vs inductive settings\\}
In the above discussed results, hashcode representations in our models are learned
% using the knowledge of data points from a test set, i.e.
in transductive setting. For inductive settings, we randomly select 25\% of the training data points for use as a pseudo-test set instead of the test set.
% , for input to \algoref{alg:opt_ref}, and we do not use any information from the test set.
% 
% we considered  settings for learning the proposed models, KLSH-NU-RF \& NLSH-NU-RF. For these models, the proposed approach of learning nearly unsupervised hashcode representations is equally applicable for inductive settings. Specifically, for additional experimental results presented in the following, we consider a random subset of a training dataset itself as a pseudo-test set; 25\% of training examples are assigned value 0 for the indicator variable $x$, and the rest of training examples are assigned value 1; using only these examples as input to \algoref{alg:opt_ref}, and not a real test set, we learn hash functions, which are then applied to the test set as well to obtain the hashcode representations for test examples.
% 
% , while the learning of hash functions purely driven from the training examples.
% 
% Since the pseudo-test set is taken as a random subset of a training set, the learned hashcode representations are purely unsupervised.
% 
% In \figref{fig:hash_sit}, we compare results from all the three choices for learning hashcode representations, supervised, transductive, and inductive.
% 
% \footnote{Of course, a Random Forest is trained in a supervised manner in all the three settings, and it is only the settings for learning hashcodes which differ.} 
% 
In \figref{fig:hash_sit}, we observe that both inductive and transductive settings are more favorable w.r.t. the supervised one, KLSH-RF, which is the baseline in this paper. F1 scores obtained from the inductive setting are on a par with the transductive settings.
% , even though the former one require no information outside a training set unlike the latter one. 
% 
It is worth noting that, in inductive settings, our model is trained on \emph{information even lesser than the baseline model} KLSH-RF, yet it obtains F1 scores significantly higher.
        
% \vspace{-1mm}
\paragraph{Neural hashing\\}
\begin{figure}[tp!]
\centering
\includegraphics[
width=\columnwidth]{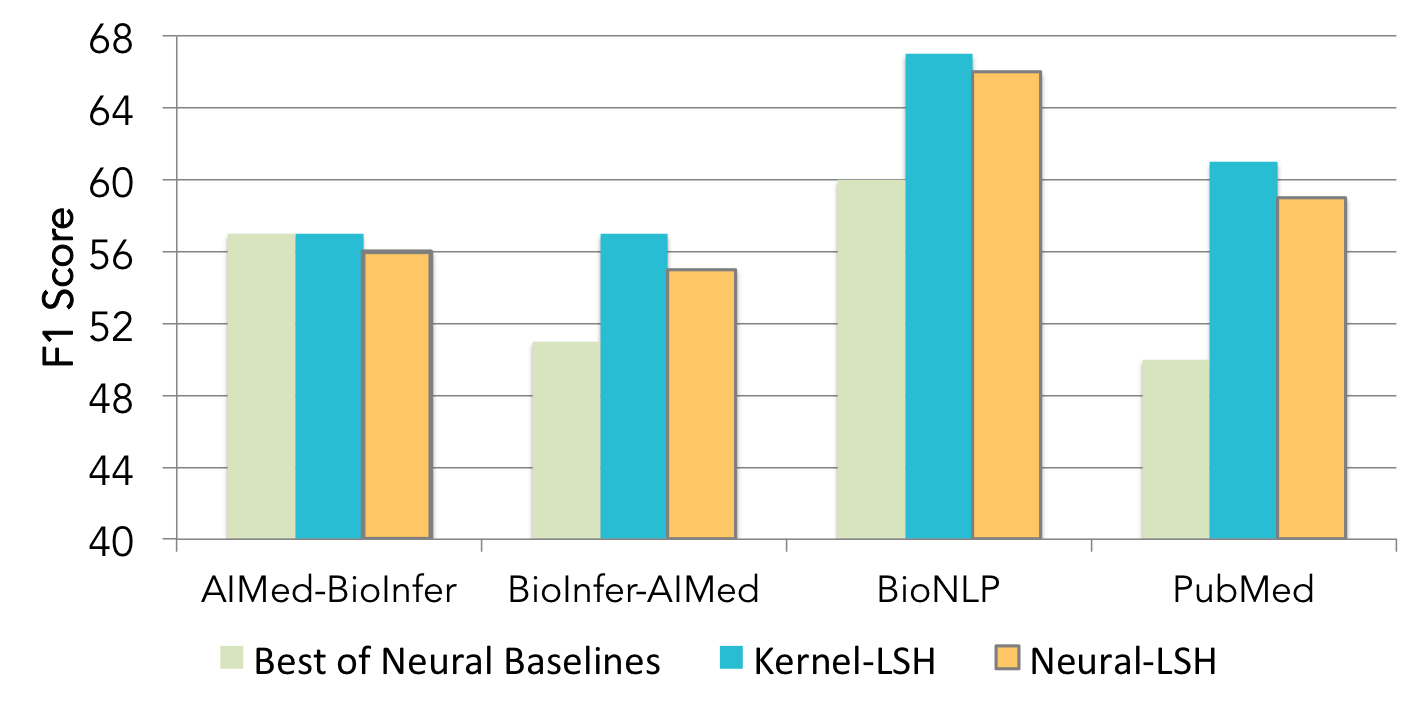}
% \vspace{-8mm}
\caption{
% \csizeten
Comparison of neural hashing w.r.t. kernel hashing, and the best of neural baselines.
}
% \vspace{-1mm}
\label{fig:neural_hash}
\end{figure}

% In reference to \tabref{tab:results_ppi} and \ref{tab:results_pubmed45_bionlp_all}, we also evaluate NLSH-NU-RF~(neural locality sensitive hashcodes from nearly unsupervised learning), and find it to be performing a little worse than its kernel based counterpart, KLSH-NU-RF, in terms of F1 scores, though it is highly competitive w.r.t. all the other baselines. 
% 
In \figref{fig:neural_hash}, we show results for neural locality sensitive hashing within our proposed framework, and observe that neural hashing is a little worse than its kernel based counterpart, however it performs significantly superior w.r.t. the best of other neural models.

% \input{related_works.tex}

% \vspace{-1mm}
\section{Conclusions}
% \vspace{-1mm}
% 
We proposed a nearly-unsupervised framework for learning of kernelized locality sensitive hashcode representations, a recent technique, that was originally supervised, which has shown state-of-the-art results for a difficult task of biomedical relation extraction. Within our proposed framework, we use the additional knowledge of test/pseudo-test data points for fine-grained optimization of hash functions so as to obtain hashcode representations generalizing across training \& test sets. Our experiment results show significant improvements in accuracy numbers w.r.t. the supervised baseline, as well as semi-supervised neural network models, for the same task of bio-medical relation extraction across four public datasets.
% 
% Further, we also demonstrated that our generic learning framework can be used for neural networks based locality sensitive hashcode representations, obtaining competitive accuracy numbers w.r.t. all the other neural baselines.
        
\section{Acknowledgments}
This work was sponsored by the DARPA Big Mechanism program (W911NF-14-1-0364). It is our pleasure to acknowledge fruitful discussions with Shushan Arakelyan, Amit Dhurandhar, Sarik Ghazarian, Palash Goyal, Karen Hambardzumyan, Hrant Khachatrian, Kevin Knight, Daniel Marcu, Dan Moyer, Kyle Reing, and Irina Rish. We are also grateful to anonymous reviewers for their valuable feedback.

\bibliography{references}
\bibliographystyle{acl_natbib}

\end{document}